\documentclass[11pt]{article}

\usepackage[preprint]{acl}

\usepackage{times}
\usepackage{latexsym}
\usepackage{listings}
\usepackage{amsmath}
\usepackage{float}
\usepackage{booktabs}
\usepackage{multirow}
\usepackage{tabularx}
\usepackage{subcaption}

\usepackage[T1]{fontenc}

\usepackage[utf8]{inputenc}

\usepackage{microtype}

\usepackage{inconsolata}

\usepackage{graphicx}

\usepackage{arydshln}
\usepackage{subcaption}

\usepackage{amsmath,amsfonts,amssymb,amsthm,commath,dsfont,nicefrac,xfrac,mathtools,mathrsfs}
\usepackage[T1]{fontenc}
\usepackage{float}
\usepackage{bm,bbm}
\usepackage[inline]{enumitem}
\usepackage{framed}
\usepackage{xspace}
\usepackage{microtype}
\usepackage[dvipsnames]{xcolor}
\usepackage[toc,page]{appendix}
\usepackage{svg}
\usepackage{booktabs}       
\usepackage{icomma}
\usepackage{hyperref}
\hypersetup{
    colorlinks=true,
    linkcolor=black,
    citecolor=blue,
    filecolor=magenta,
    urlcolor=blue
}
\usepackage{soul}
\usepackage{cleveref}

\usepackage{caption}
\usepackage{subcaption}
\Crefname{equation}{Eq.\!}{Eqs.\!}
\Crefname{subtable}{Table}{Tables}

\theoremstyle{plain}

\theoremstyle{plain}

\theoremstyle{plain}

\usepackage{graphicx}
\usepackage{wrapfig}

\title{Your Reasoning Benchmark May Not Test Reasoning: Revealing Perception Bottleneck in Abstract Reasoning Benchmarks}

\author{Xinhe Wang \\
  Carnegie Mellon University \\
  \texttt{xinhew@andrew.cmu.edu} \\\And
  Jin Huang \\
  University of Michigan \\
  \texttt{huangjin@umich.edu} \\\And
  Xingjian Zhang \\
  University of Michigan \\
  \texttt{jimmyzxj@umich.edu} \\
  \AND
  Tianhao Wang \\
  University of California San Diego \\
  \texttt{tianhaowang@ucsd.edu} \\\And
  Jiaqi W. Ma \\
  University of Illinois Urbana-Champaign \\
  \texttt{jiaqima@illinois.edu} \\}

\begin{document}
\maketitle

\begin{abstract}
Reasoning benchmarks such as the Abstraction and Reasoning Corpus (ARC) and ARC-AGI are widely used to assess progress in artificial intelligence and are often interpreted as probes of core, so-called ``fluid'' reasoning abilities. Despite their apparent simplicity for humans, these tasks remain challenging for frontier vision-language models (VLMs), a gap commonly attributed to deficiencies in machine reasoning. We challenge this interpretation and hypothesize that the gap arises primarily from limitations in visual perception rather than from shortcomings in inductive reasoning.

To verify this hypothesis, we introduce a two-stage experimental pipeline that explicitly separates perception and reasoning. In the perception stage, each image is independently converted into a natural-language description, while in the reasoning stage a model induces and applies rules using these descriptions. This design prevents leakage of cross-image inductive signals and isolates reasoning from perception bottlenecks. Across three ARC-style datasets, Mini-ARC, ACRE, and Bongard-LOGO, we show that the perception capability is the dominant factor underlying the observed performance gap by comparing the two-stage pipeline with against standard end-to-end one-stage evaluation. Manual inspection of reasoning traces in the VLM outputs further reveals that approximately 80 percent of model failures stem from perception errors. 
Together, these results demonstrate that ARC-style benchmarks conflate perceptual and reasoning challenges and that observed performance gaps may overstate deficiencies in machine reasoning. Our findings underscore the need for evaluation protocols that disentangle perception from reasoning when assessing progress in machine intelligence.
\end{abstract}
\section{Introduction}
Reasoning capability has increasingly become a central criterion for evaluating the progress of frontier artificial intelligence (AI) models~\citep{deepmind_gemini2025,openai_o3_video2024}. Correspondingly, a wide range of reasoning benchmarks have emerged to assess different dimensions of reasoning performance~\citep{arcprize_arcagi1,yue2025mmmu,chen2021evaluating,rein2024gpqa}. Among these, the Abstraction and Reasoning Corpus (ARC)~\citep{chollet2019arc} and its successor, ARC-AGI ~\citep{arcprize_arcagi1,chollet2025arc}, have attracted particular attention, as they are widely viewed as tests of core reasoning abilities and, in some views, as indicators of progress toward general intelligence~\citep{chollet2024technical}. Reflecting their influence, frontier AI labs, including OpenAI and Google, routinely highlight their models’ performance on ARC as a key indicator of advances in general reasoning capabilities~\citep{deepmind_gemini2025,arcprize_arcagi1,openai_o3_video2024}.

\begin{figure*}[ht]
  \centering
  \begin{minipage}{0.45\textwidth}
    \centering
    \includegraphics[width=\textwidth]{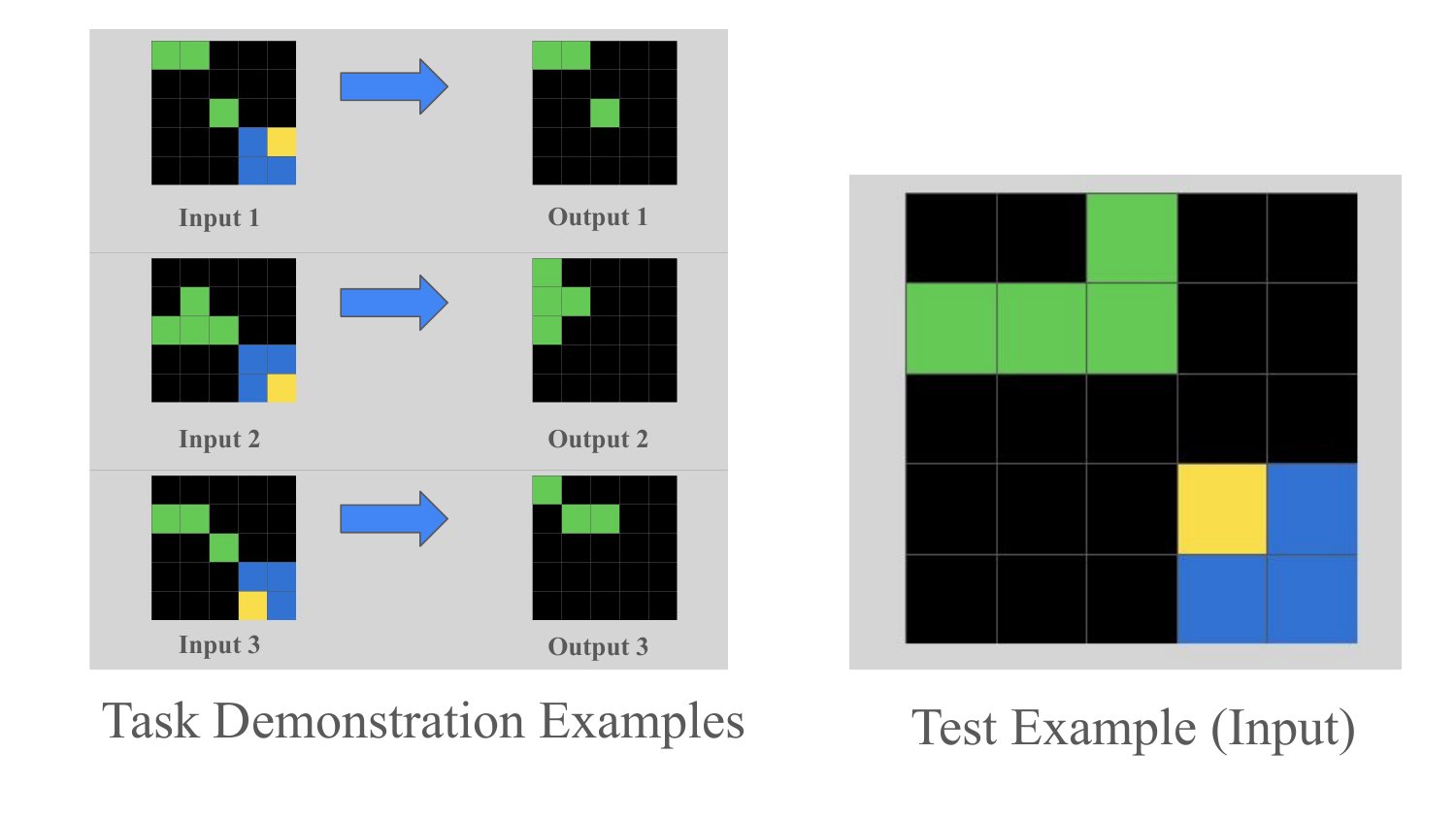}
    \subcaption{An example ARC problem shown in 2-D visually.}
    \label{fig:visual-arc-example}
  \end{minipage}
  \hfill
  \begin{minipage}{0.47\textwidth}
    \centering
\begin{lstlisting}[basicstyle=\ttfamily\scriptsize, breaklines=true, breakindent=0pt]
Task Demonstration Examples:

Input: [[0, 0, 0, 0, 0], [3, 3, 0, 0, 0], [0, 0, 3, 0, 0], [0, 0, 0, 1, 1], [0, 0, 0, 4, 1]]
Output: [[3, 0, 0, 0, 0], [0, 3, 3, 0, 0], [0, 0, 0, 0, 0], [0, 0, 0, 0, 0], [0, 0, 0, 0, 0]]

(Omitting two demonstration examples.)

Test Example:

Input: [[0, 0, 3, 0, 0], [3, 3, 3, 0, 0], [0, 0, 0, 0, 0], [0, 0, 0, 4, 1], [0, 0, 0, 1, 1]]
    \end{lstlisting}
\subcaption{The same ARC problem shown in a serialized format.}
\label{fig:serial-arc-example}
  \end{minipage}
  \caption{An example ARC problem. One is asked to induce a common rule from the input-output pairs in the task demonstration examples, and then apply this rule to the test input to generate the test output. The difficulty of this task to human players critically depends on how the problem is presented.}
  \label{fig:arc-example}

\end{figure*}

The core design principle of ARC and ARC-AGI is to evaluate ``\emph{fluid} intelligence (the ability to reason, solve novel problems, and adapt to new situations) rather than \emph{crystallized} intelligence, which relies on accumulated knowledge and skills,''~\citep{arcprize_arcagi1} where the former is considered a core reasoning capability at which current AI models still fall short. Concretely, ARC and ARC-AGI take the form of grid-based puzzles, as shown in \Cref{fig:visual-arc-example}. Each problem consists of several input-output demonstration pairs and a test input. The solver--human or AI--must induce a common rule from the demonstrations and then apply it to the test input. In the specific example in \Cref{fig:visual-arc-example}, the correct solution requires recognizing the pattern that the $2\times2$ block on the bottom-right corner encodes the rotation rule for the 3x3 green object in the upper-left, and the position of the yellow pixel indicates degrees of rotation. The solver must then apply this inferred rule to the test input to generate the correct test output. 

Empirically, these problems are found to be easy for humans yet surprisingly hard for even state-of-the-art frontier AI models\footnote{Throughout this paper, we focus on vision-language models (VLMs) unless stated otherwise, as most frontier models are evaluated on these benchmarks in VLM form. We note that other kinds of models specialized for ARC-style tasks fall outside the scope of our discussion.}~\citep{arcprize_arcagi1}. A common belief is that this persistent gap between AI and human reflects a fundamental advantage of humans' reasoning capabilities~\citep{chollet2024technical}. In this work, we critically examine this belief, and \emph{hypothesize} that \ul{the gap arises primarily because ARC problems particularly favor humans' innate visual \textbf{perception}, rather than reflecting a genuine difference in the \textbf{reasoning} capability (or ``fluid intelligence'').} This hypothesis is motivated by the observation that the difficulty of the ARC problem to human players depends strongly on how the problem is presented. As shown in \Cref{fig:arc-example}, presenting the same problem in a serialized format (\Cref{fig:serial-arc-example}) makes it far more difficult for humans to solve than when shown in its original 2-D visual format (\Cref{fig:visual-arc-example}).

Our hypothesis involves the explicit separation of \textbf{perception} and \textbf{reasoning} capabilities required to solve ARC-style tasks. Conceptually, perception refers to the ability to \emph{recognize meaningful objects from raw visual inputs}, whereas reasoning refers to the ability to \emph{induce patterns among the recognized objects in demonstrations}. A key challenge in verifying our hypothesis lies in the fact that the success of perception is a prerequisite for the success of reasoning in solving an ARC-style problem, which makes it difficult to have a clean measure and comparison of the perception and reasoning capabilities of a given model. 

To address this challenge and verify our hypothesis, we design a two-stage experimental pipeline. The first stage (\emph{perception stage}) transforms the raw image inputs into natural language descriptions. Crucially, this transformation is \emph{applied to each images in isolation}, ensuring that no cross-image inductive signals are leaked during this stage. This atomistic approach guarantees that inductive reasoning occurs exclusively in the second stage (\emph{reasoning stage}), where an AI model is tasked with solving the problem using the natural language descriptions obtained from the perception stage. The natural-language-based representation is supposed to alleviate the perception challenge for the model while preserving the inductive structure of the problem. This pipeline allows us to isolate the model's inductive reasoning performance from its perceptual bottlenecks, providing a clearer picture of where the ``reasoning gap'' truly resides.

We conduct experiments on three ARC-style visual reasoning datasets, Mini-ARC~\citep{kim2022playgrounds}, ACRE~\citep{zhang2021acre}, and Bongard-LOGO~\citep{nie2020bongard}. We compare the performance of vision-language models (VLMs) under both the standard end-to-end one-stage setting and the proposed two-stage pipeline. Our empirical results provide three key findings that support our hypothesis. First, we demonstrate that the two-stage pipeline, in which a dedicated perception stage transforms the raw image inputs into natural language descriptions, significantly outperforms the end-to-end one-stage application of the same VLM. Second, we find that a hybrid two-stage pipeline combining a strong VLM for the perception stage with a weaker VLM for the reasoning stage yields performance close to that of an end-to-end strong VLM, while substantially outperforming an end-to-end weak VLM. This result suggests that perceptual capability, rather than reasoning strength, is the primary bottleneck in these tasks. Finally, through manual inspection of model reasoning traces, we observe that approximately 80\% of failure cases stem from perception errors, i.e., failing to properly identify visual objects. Furthermore, the majority of the performance gains achieved by the two-stage pipeline can be attributed to a reduction in such perception errors.

In summary, our study reveals an important perception bottleneck in an influential class of reasoning benchmarks, abstract reasoning benchmarks, through carefully controlled experiments. Our results suggest that performance gaps on ARC-style tasks may conflate limitations in visual perception with deficiencies in inductive reasoning. This finding calls for caution in interpreting these benchmarks as direct measures of reasoning or fluid intelligence in frontier AI models, and highlights the importance of disentangling perceptual and reasoning components when evaluating progress in machine reasoning.
\section{Related Work}

\paragraph{Reasoning Benchmarks.}
Reasoning benchmarks can be categorized by (i) the input modality and (ii) how much they rely on external knowledge versus in-context rule induction. Text-only benchmarks probe reasoning entirely in language, spanning broad knowledge and multi-disciplinary question answering (e.g., \textsc{MMLU}~\citep{hendrycks2020measuring}), multi-step mathematical problem solving (e.g., \textsc{GSM8K}~\citep{cobbe2021training}), functional code generation (e.g., \textsc{HumanEval}~\citep{chen2021evaluating}), and harder challenge suites designed to reduce superficial heuristics (e.g., \textsc{GPQA}~\citep{rein2024gpqa}, \textsc{Big-Bench Hard}~\citep{suzgun2023challenging}). Vision-language benchmarks explicitly couple perception (e.g., reading, grounding, extracting structure from images) with knowledge-based reasoning, including scientific and mathematical reasoning over diagrams (e.g., \textsc{ScienceQA}~\citep{lu2022learn}, \textsc{MathVista}~\citep{lu2023mathvista}), text-rich visual understanding (e.g., \textsc{DocVQA}~\citep{mathew2021docvqa}, \textsc{TextVQA}~\citep{singh2019towards}), and comprehensive multi-domain suites (e.g., \textsc{MMMU}~\citep{yue2024mmmu}, \textsc{MMBench}~\citep{liu2024mmbench}). Finally, knowledge-light visual abstraction benchmarks aim to minimize human priors by emphasizing pattern recognition, abstraction, and generalization from minimal visual examples (e.g., ARC~\citep{arcprize_arcagi1}, ACRE~\citep{zhang2021acre} and \textsc{Bongard-LOGO}~\citep{nie2020bongard}).  These knowledge-light abstraction benchmarks are the primary focus of our work: although they are widely viewed as reasoning-centric tests that require minimal perceptual effort, we demonstrate that their performance can be strongly limited by perceptual bottlenecks, and that substantial gains can arise from improving perception rather than advancing reasoning.

\paragraph{State-of-the-Art Performance on the ARC Benchmarks.} Recent progress on ARC/ARC-AGI can be broadly grouped into (i) general-purpose foundation models, including LLMs that serialize grids into text VLMs that operate directly on the visual grid representation; and (ii) tailored ARC solvers that introduce task-specific search, program synthesis, or specialized architectures. 
For general-purpose foundation models, GPT-5.2 Pro (High) and Gemini 3 Pro (Deep Think) have established a new ceiling, achieving scores of 54.2\% and 45.1\% respectively in ARC-AGI-2 primarily due to their extended thinking ability~\citep{arcprize_leaderboard}.
This shift towards scaling test-time computation has crystallized into a new paradigm named ``refinement loops,'' where systems iteratively optimize solutions against feedback rather than relying on single-shot inference.

Among tailored approaches, test-time training (TTT) has emerged as a strong general mechanism for few-shot adaptation on ARC, substantially improving over fine-tuned baselines and often combining well with other solver components~\citep{li2024combining,pmlr-v267-akyurek25a}.
Alternative neural approaches include the Hierarchical Reasoning Model (HRM) and its recursive variants, which utilize small recurrent architectures~\citep{wang2025hierarchical,jolicoeur2025less}, and masked diffusion models, which refine the grid globally to capture structural constraints~\citep{franzen2025architects}. Additionally, CompressARC explores test-time learning by minimizing description length on the target puzzle~\citep{liao2025arc}.
Finally, a very recent concurrent work argues that ``ARC is a vision problem''~\citep{hu2025arc}, reframing ARC as image-to-image translation with a ViT-style backbone and test-time adaptation, which gives an example of a vision-centric route to ARC performance. 
While this recent trend suggests that enhancing perception boosts performance, our primary goal is to challenge the interpretation of ARC performance as a proxy for general intelligence. Therefore, in this paper, we focus specifically on general-purpose VLMs, aiming to understand how much of their ARC performance is limited by perception rather than reasoning. 

\paragraph{Limitations of the ARC Benchmarks.}  The research community has suggested several limitations that complicate treating ARC scores as a clean measure of general reasoning. First, strong performance can result from increased compute, either through larger training budgets or extensive test-time training or searching~\citep{mitchell2024openai}. Therefore, gains in accuracy may reflect better fitting to the ARC task distribution rather than improved reasoning. Second, the small hidden evaluation set increases the risk of implicit overfitting to benchmark-specific patterns~\citep{chollet2024technical}. Third, ARC tasks are inherently \emph{visual}, yet many recent solutions are language-centric~\citep{li2024combining, pmlr-v267-akyurek25a}. A concurrent work shows that reframing ARC as an image-to-image translation problem and applying standard vision architectures can achieve near-human performance on ARC-1~\citep{hu2025arc}. This result suggests that high ARC scores can be obtained by improving visual representations, without necessarily requiring more general reasoning mechanisms.
Our work builds on these studies and examines how perception and reasoning interact when using VLMs to solving ARC tasks.

\section{Verifying the Hypothesis with a Two-Stage Pipeline}

In this section, we formally introduce the design of a two-stage pipeline to verify the hypothesis that the performance of existing VLMs is constrained more by perception than by reasoning.

\subsection{Abstract Reasoning Tasks}
\begin{figure}
\vspace{-6mm}
    \centering
    \includegraphics[width=1\linewidth]{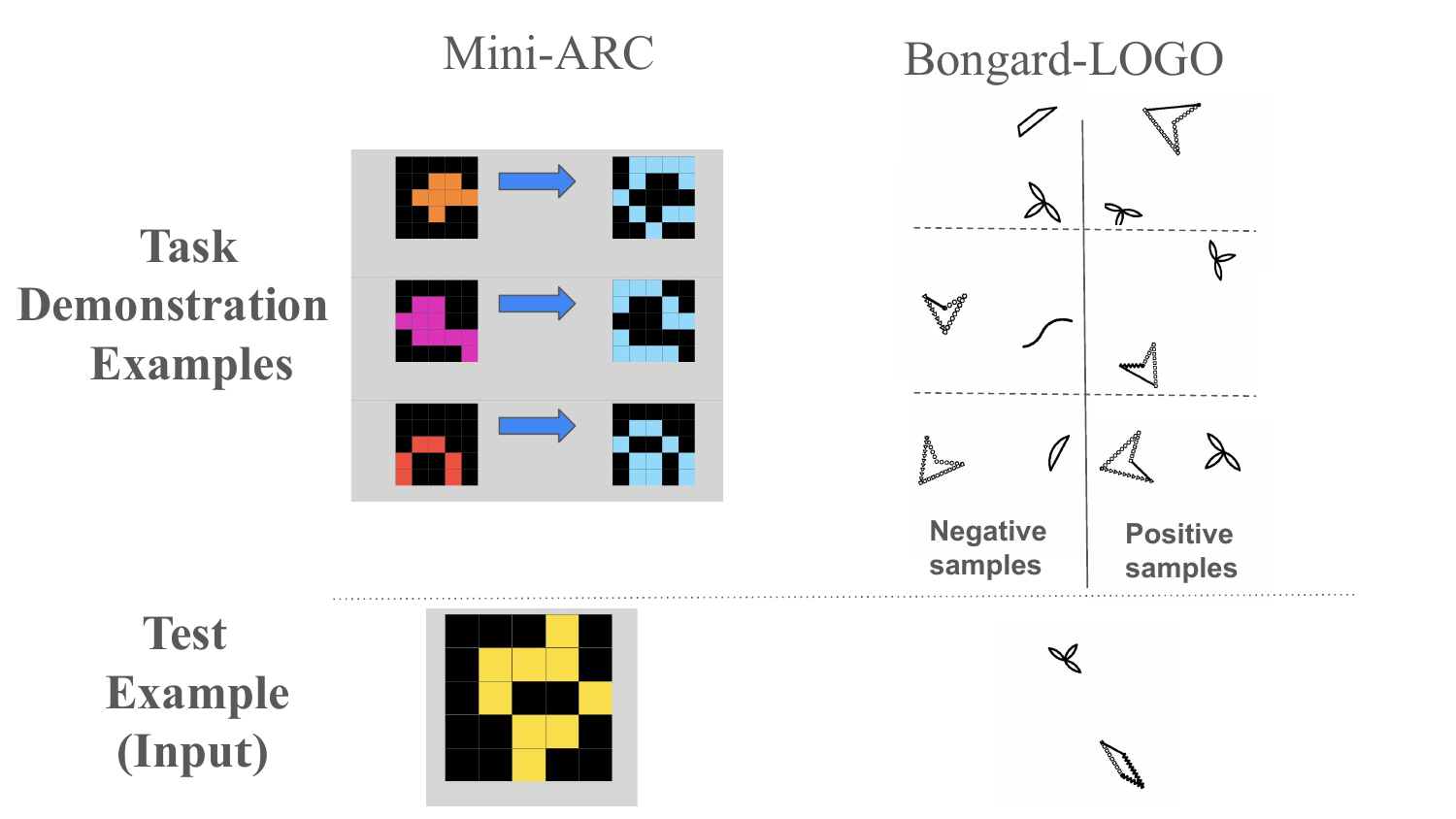}
    \caption{Example tasks of two ARC-style benchmarks, Mini-ARC~\citep{kim2022playgrounds} and Bongard-LOGO~\citep{nie2020bongard}. The task in Mini-ARC maps image inputs to image outputs. The task in Bongard-LOGO maps image inputs to binary outputs (positive or negative). In both cases, visual objects that are immediately recognizable by humans serve as the critical basis for the latent mapping rules.}
    \label{fig:benchmark-examples}
    \vspace{-5mm}
\end{figure}

We start by formalizing the (visual) abstract reasoning tasks considered in this study. Let $\mathcal{T}$ denote a set of tasks, where each task $T \in \mathcal{T}$ consists of a small number $n$ (typically $n \le 10$) of input-output pairs $(x_i, y_i) \in \mathcal{X}\times\mathcal{Y}$ serving as demonstration examples, together with a test example for which only the input ($x_{n+1}$) is observed. Formally,
\[
T = \left\{ (x_1, y_1), \ldots, (x_n, y_n), \; x_{n+1} \right\}.
\]
The goal of the task is to predict the corresponding output $y_{n+1}$ by inductively inferring the underlying relationship between inputs and outputs from the demonstration examples. Figure~\ref{fig:benchmark-examples} illustrates example tasks from two ARC-style benchmarks, where each task involves images as inputs and, in some cases, as both inputs and outputs. A shared characteristic of these benchmarks is that the \ul{images contain objects that are are immediately and unambiguously recognizable to humans, which serve as the critical basis for the latent mapping rules}.

When evaluated on these benchmarks, a frontier VLM model is typically used as a mapping $f: \mathcal{T} \rightarrow \mathcal{Y}$ that directly predicts $y_{n+1}$ with $f(T)$, given the task $T$ with raw image inputs. However, the model can significantly underperform humans in recognizing the objects visually salient to humans. In this end-to-end evaluation paradigm, it is difficult to quantify the extent to which the model performance is limited by the model’s perception capabilities compared to its reasoning capabilities.

\subsection{Quantifying the Perception Bottleneck with a Two-Stage Pipeline}

To verify our hypothesis and quantify the perception bottleneck of VLMs, we evaluate the models with a two-stage pipeline that explicitly separates \textbf{perception} and \textbf{reasoning}. In the first stage (the perception stage), images are transformed into natural language descriptions. In the second stage (the reasoning stage), the model takes these descriptions to enrich the task representation $T$, and predicts the output $y_{n+1}$ on the test example.

\paragraph{Design Principles of the Perception Stage.} The perception stage is designed according to two key principles:
\begin{enumerate}
    \item \textbf{No cross-image inductive signal leakage.} The transformation is applied independently to each image in isolation, ensuring that it reduces perceptual difficulty without introducing inductive cues that could alter the intrinsic reasoning difficulty of the task.
    \item \textbf{Generic human perceptual priors.} The transformation incorporates generic human perceptual priors, such as the identification of objects and recognition of their colors or shapes. This choice directly reflects our hypothesis that the human–AI performance gap is largely driven by the fact that these benchmarks favor human's innate visual perception.
\end{enumerate}

\paragraph{Formal Description of the Two-Stage Pipeline.} Formally, for each benchmark, we construct two \emph{uniform} transformations, $g_{\mathcal{X}}: \mathcal{X} \rightarrow \widetilde{\mathcal{X}}$ and $g_{\mathcal{Y}}: \mathcal{Y} \rightarrow \widetilde{\mathcal{Y}}$, which are respectively applied to the input and output spaces, consistently across all tasks within the benchmark. In practice, these transformations are implemented by prompting a VLM model to convert each image into a corresponding natural language description, with prompts that explicitly instruct the model to attend to features aligned with generic human visual priors. When the output is not an image, the transformation $g_{\mathcal{Y}}$ is defined as the identity mapping.
With these transformations, a task $T = \left\{ (x_1, y_1), \ldots, (x_n, y_n), \; x_{n+1} \right\}$ will be enriched as
\begin{align*}
    \widetilde{T} = \Big\{ & (x_1, g_{\mathcal{X}}(x_1), y_1, g_{\mathcal{Y}}(y_1)), \ldots, \\
    & (x_n, g_{\mathcal{X}}(x_n), y_n, g_{\mathcal{Y}}(y_n)), \; \\
    & (x_{n+1}, g_{\mathcal{X}}(x_{n+1})) \Big\} \in \widetilde{\mathcal{T}},
\end{align*}
which will be further fed into a VLM, $h:\widetilde{\mathcal{T}} \rightarrow \mathcal{Y}$, during the reasoning stage to complete the prediction $h(\widetilde{T})$.

\subsection{Two Evaluation Settings}
\label{sec:eva-settings}
We consider two evaluation settings that differ in how VLMs are instantiated within the two-stage pipeline.

\paragraph{Setting 1: Same-Model Perception.} In the first setting, we use the same VLM for the transformations $g_{\bullet}$, the reasoning stage $h$, as well as the baseline end-to-end one-stage predictor $f$. In this case, we expect the two-stage prediction $h(\widetilde{T})$ to outperform the one-stage prediction $f(T)$, as the perception stage explicitly incorporates additional human perceptual priors. \textbf{Importantly}, \ul{any performance improvement in this comparison can be attributed solely to the mitigation of the perception bottleneck}, since the two-stage pipeline does not reduce the inductive reasoning difficulty by design\footnote{In fact, the two-stage pipeline may slightly increase the inductive reasoning difficulty since the model must perform induction with more information.}.

\paragraph{Setting 2: Stronger-Model Perception.} In the second setting, we use a stronger VLM for the transformations $g_{\bullet}$, and a weaker VLM for the reasoning stage $h$. In this case, we compare this hybrid two-stage pipeline against end-to-end one-stage predictions using the strong and weak models, denoted by $f_S(T)$ and $f_W(T)$, respectively. First, we expect $h(\widetilde{T})$ to substantially outperform $f_W(T)$, since the perception stage is further enhanced. Second, if the performance of $h(\widetilde{T})$ approaches that of $f_S(T)$, this suggests that \ul{the performance gap between the weak and strong models is driven primarily by differences in perception capability rather than reasoning capability}.

\subsection{Fine-Grained Error Attribution with Four Categories}\label{sec:error-attribution-approach}

\begin{figure}[ht]
\vspace{-6mm}
    \centering
    \includegraphics[width=0.85\linewidth]{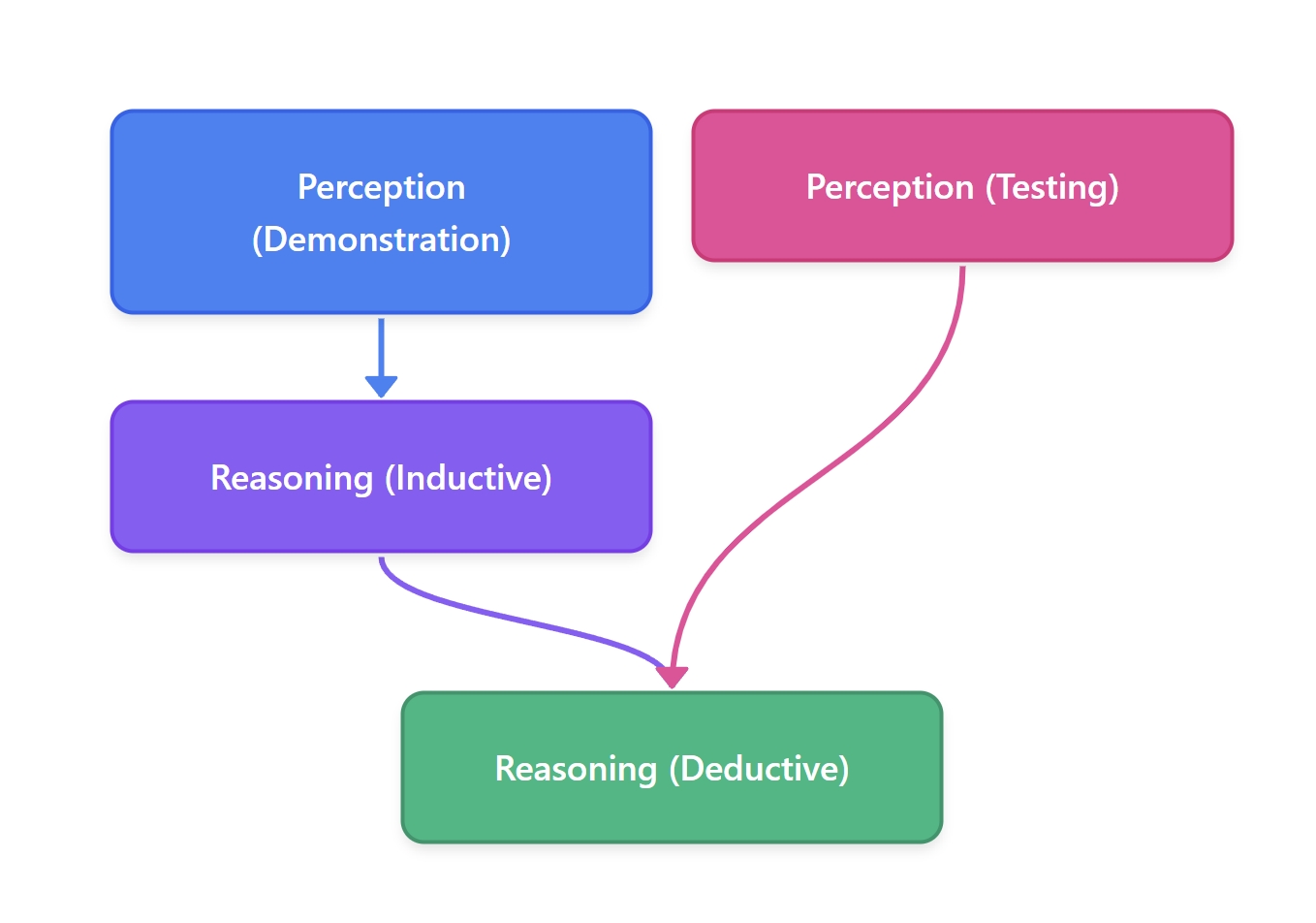}
    \caption{Dependency graph illustrating the four-step task-solving process. Errors in earlier stages propagate to subsequent stages, affecting final predictions.}
    \vspace{-4mm}
    \label{fig:dependency-graph}
\vspace{-3mm}
\end{figure}

\paragraph{Conceptual Decomposition of the Task-Solving Process.} Given the similar task structure of the abstract reasoning benchmarks, we conceptually decompose the task-solving process into four steps:
\begin{enumerate}
    \item \textbf{Perception (Demonstration).}  
    The system must correctly perceive each input pair $(x_i, y_i)$.
    \item \textbf{Reasoning (Inductive).}  
    From the correctly perceived demonstrations, the system must infer the underlying latent mapping rule that governs the task.
    \item \textbf{Perception (Testing).}  
    The system must also accurately perceive the test input.
    \item \textbf{Reasoning (Deductive).}  
    Given the inferred rule, the system applies it to the test input to make the correct prediction.
\end{enumerate}
This decomposition distinguishes steps (1) and (3), which primarily depend on perception capabilities, from steps (2) and (4), which primarily depend on reasoning capabilities. It is worth noting that the four steps are not independent. For example, an error in the step 1 will propagate to and cause failures in steps (2) and (4). An illustration of the dependency of these steps is shown in~\ref{fig:dependency-graph}.
\paragraph{Error Attribution with Four Categories.} We observe that the failure cases in our experiments can almost always be attributed to one of four error categories corresponding to the four steps above. In practice, this error attribution can be done by manually inspecting the reasoning traces in the VLM outputs. This procedure is applicable to both one-stage and two-stage predictions. To enable a fair comparison between one-stage and two-stage predictions on the same benchmark, we assign the prediction on each task to one of five categories: the four error categories and a \emph{Correct} category, ensuring that total counts match across the two prediction settings. Given the largely sequential dependency among the four steps, errors are attributed to the earliest step at which a failure occurs. In Section~\ref{sec:error-attribution}, we conduct large-scale error attribution across multiple benchmarks for both one-stage and two-stage predictions, and demonstrate that the performance gains of the two-stage pipeline over the one-stage baseline arise almost exclusively from a reduction in perception errors, i.e., errors in steps (1) and (3).

\section{Experiments}
\label{sec:experiments}
\subsection{Datasets}
We conduct experiments on three ARC-style visual abstract reasoning benchmarks. These datasets are designed to assess compositional and inductive reasoning from visual inputs, making them well suited for studying the interplay between perception and reasoning in VLMs.

\paragraph{Mini-ARC.} Mini-ARC~\citep{kim2022playgrounds} is a reduced-scale variant of ARC with fewer colors, smaller grids, and simplified transformations. It maintains the same  format as ARC but with lower perceptual complexity, making it useful for isolating reasoning performance under easier perception.

\paragraph{ACRE (Abstract Causal REasoning Beyond Covariation).} ACRE~\citep{zhang2021acre} is a symbolic visual reasoning benchmark designed for causal induction. Each task is formulated as a classification problem conditioned on demonstrations. Specifically, a task consists of six demonstrations followed by four prediction queries. The demonstrations illustrate the presence or absence of certain objects and the corresponding state of a pink board, which may blink (activated), remains unlit (deactivated), or be underdetermined. Each query presents a new configuration of objects, and the objective is to predict the status of the pink board.

\paragraph{Bongard-LOGO.} Bongard-LOGO~\citep{nie2020bongard} is a visual reasoning benchmark inspired by the classic Bongard problems, adapted to the domain of programmatic graphics. Each task presents two sets of images: positive examples and negative examples. The images are generated from programs in the LOGO language, with the positive set sharing an underlying semantic concept that is absent from the negative set. Given these examples, the model must infer the underlying concept and correctly classify novel test instances as belonging to the positive or negative set.
\vspace{-1mm}
\subsection{Setting 1: Same-Model Perception}

We first report the experiments corresponding to the first evaluation setting (Same-Model Perception) described in Section~\ref{sec:eva-settings}, where we compare the two-stage pipeline and the one-stage baseline using the same VLM. We use GPT-4o for Mini-ARC and Bongard-LOGO, while using LLaVA-1.5~\citep{liu2024improved} for ACRE. Table~\ref{tab:exp-setup} summarizes concrete settings for each dataset.

\begin{table}[ht]
\vspace{-4mm}
\centering
\footnotesize 
\renewcommand{\arraystretch}{1.2} 
\setlength{\tabcolsep}{2.8pt} 
\caption{Experimental setups for each dataset in Setting 1 (Same-Model Perception).
We have two configurations (a) and (b) for each dataset.
Stage: ``S'' refers to standard one-stage pipeline; ``P+R'' refers to two-stage pipeline with separate perception (``P'') and reasoning (``R'').}
\label{tab:exp-setup}
\begin{tabularx}{\linewidth}{@{}c@{\hspace{2pt}}c@{\hspace{2pt}}c@{\hspace{2pt}}c@{\hspace{2pt}}>{\raggedright\arraybackslash}X@{}}
\toprule
\textbf{ID} & \textbf{Dataset} & \textbf{Config} & \textbf{Stage} & \textbf{Model} \\
\midrule

\multirow{2}{*}{1} & \multirow{2}{*}{Mini-ARC}
  & (a) & S   & GPT-4o \\
  &     & (b) & P+R & GPT-4o (P) + GPT-4o (R) \\
\cdashline{1-5}

\multirow{2}{*}{2} & \multirow{2}{*}{Bongard-LOGO}
  & (a) & S   & GPT-4o \\
  &     & (b) & P+R & GPT-4o (P) + GPT-4o (R) \\
\cdashline{1-5}

\multirow{2}{*}{3} & \multirow{2}{*}{ACRE}
  & (a) & S   & LLaVA-1.5 \\
  &     & (b) & P+R & LLaVA-1.5 (P) + LLaVA-1.5 (R) \\

\bottomrule
\end{tabularx}
\vspace{-4mm}
\end{table}

\paragraph{Results.} We report the experiment results in Table~\ref{tab:overall-results}. Across three datasets, enhancing perception through natural language descriptions consistently improves success rates by 11–13 percentage points. Notably, on Mini-ARC this corresponds to a $2.5\times$ relative improvement, increasing performance from 8.05\% to 20.13\%.
These results are consistent with our expectation for Setting 1 described in Section~\ref{sec:eva-settings}, which supports our hypothesis that perception plays a more important role for the success in these tasks.

\begin{table}[ht]
\vspace{-2mm}
\centering
\small
\setlength{\tabcolsep}{3pt}
\caption{
Success rates (\%) in Setting 1 (Same-Model Perception).
The columns (a) and (b) correspond to the configurations defined in Table~\ref{tab:exp-setup}. $\Delta$ refers to the absolute improvement from (a) to (b) in percentage points.
}
\label{tab:overall-results}
\begin{tabular}{c l c c c}
\toprule
\textbf{ID} & \textbf{Dataset} & \textbf{(a)} & \textbf{(b)} & $\boldsymbol{\Delta}$ \\
\midrule

1 & Mini-ARC     & 8.05  & 20.13 & +12.08 \\
2 & Bongard-LOGO & 62.00 & 73.00 & +11.00 \\
3 & ACRE         & 22.00 & 34.50 & +12.50 \\

\bottomrule
\end{tabular}
\vspace{-3mm}
\end{table}

\subsection{Setting 2: Stronger-Model Perception}

We further conduct experiments corresponding to the second evaluation setting (Stronger-Model Perception) described in Section~\ref{sec:eva-settings}, where we use a stronger VLM for the perception stage in the two-stage pipeline. We conduct these experiments on the three datasets mentioned above, where we replace the weaker model used in the perception stage with a stronger model. We also compare against standard one-stage pipelines that use a weak model or a strong model. For Mini-ARC, we consider two different strong models: o1 and Claude-Sonnet-4.5. The concrete configurations are summarized in Table~\ref{tab:stronger-perception-setup}.

\begin{table}[ht]
\vspace{-3mm}
\centering
\footnotesize
\setlength{\tabcolsep}{3pt}
\caption{
Experimental setups in Setting 2 (Stronger-Model Perception).
Configurations (a) and (b) are identical to those in Table~\ref{tab:exp-setup}.
Configurations (c) and (d) introduce stronger models either in the perception stage or in a unified one-stage pipeline.
}
\label{tab:stronger-perception-setup}
\begin{tabularx}{\linewidth}{@{}c c c >{\raggedright\arraybackslash}X@{}}
\toprule
\textbf{ID} & \textbf{Dataset} & \textbf{Config} & \textbf{Model(s)} \\
\midrule

\multirow{6}{*}{1} 
& \multirow{6}{*}{Mini-ARC}
& (a) & GPT-4o (S) \\
& & (b) & GPT-4o (P) + GPT-4o (R) \\
& & (c1) & o1 (P) + GPT-4o (R) \\
& & (c2) & Claude-Sonnet-4.5 (P) + GPT-4o (R) \\
& & (d1) & o1 (S) \\
& & (d2) & Claude-Sonnet-4.5 (S) \\
\cdashline{1-4}

\multirow{4}{*}{2}
& \multirow{4}{*}{Bongard-LOGO}
& (a) & GPT-4o (S) \\
& & (b) & GPT-4o (P) + GPT-4o (R) \\
& & (c) & o1 (P) + GPT-4o (R) \\
& & (d) & o1 (S) \\
\cdashline{1-4}

\multirow{4}{*}{3}
& \multirow{4}{*}{ACRE}
& (a) & LLaVA-1.5 (S) \\
& & (b) & LLaVA-1.5 (P) + LLaVA-1.5 (R) \\
& & (c) & GPT-4o (P) + LLaVA-1.5 (R) \\
& & (d) & GPT-4o (S) \\
\bottomrule
\end{tabularx}
\vspace{-4mm}
\end{table}

 \begin{table}[ht]
\centering
\small
\setlength{\tabcolsep}{4pt}
\caption{
Success rates (\%) under Setting~2 (Stronger-Model Perception).
Configurations (a)--(d) correspond to those defined in Table~\ref{tab:stronger-perception-setup}.
}
\label{tab:setting2-results}
\begin{tabular}{l c c c c}
\toprule
\textbf{Dataset} & \textbf{(a)} & \textbf{(b)} & \textbf{(c*)} & \textbf{(d*)} \\
\midrule
\multirow{2}{*}{Mini-ARC}      & \multirow{2}{*}{8.05}  & \multirow{2}{*}{20.13} & (c1) 31.54 & (d1) 52.03 \\
 & & & (c2) 32.89 & (d2) 34.22 \\
Bongard-LOGO         & 62.00 & 73.00 & 80.00 & 78.00 \\
ACRE                 & 22.00 & 34.50 & 82.50 & 93.00 \\
\bottomrule
\vspace{-10mm}
\end{tabular}
\end{table}

\begin{table*}[ht]
\vspace{-8mm}
\centering
\footnotesize
\setlength{\tabcolsep}{5pt}
\renewcommand{\arraystretch}{1.05}
\caption{Error attribution across datasets and experiment configurations. Regarding the settings, for example, ``1(a)'' refers to the configuration (a) on Mini-ARC, while ``3(c)'' refers to the configuration (c) on ACRE.}
\label{tab:error-attrib-main}
\vspace{-1mm}

\begin{minipage}{0.48\linewidth}
\centering
\subcaption{Mini-ARC}
\begin{tabularx}{\linewidth}{@{}lrr@{}}
\toprule
Setting & 1(a) & 1(b) \\
\midrule
\textbf{Total Errors} & 44  & 37  \\
Perception (Demo) & 38 (86.4\%) & 22 (59.5\%) \\
Reasoning (Inductive) & 4 (9.1\%) & 9 (24.3\%) \\
Perception (Test) & 1 (2.3\%) & 2 (5.4\%) \\
Reasoning (Deductive) & 1 (2.3\%) & 4 (10.8\%) \\
\bottomrule
\end{tabularx}
\end{minipage}
\hfill
\begin{minipage}{0.48\linewidth}
\centering
\subcaption{Bongard-LOGO}
\begin{tabularx}{\linewidth}{@{}lrr@{}}
\toprule
Setting & 2(a) & a(b) \\
\midrule
\textbf{Total Errors} & 38  & 27 \\
Perception (Demo) & 25 (65.8\%) & 10 (37.0\%) \\
Reasoning (Inductive) & 5 (13.2\%) & 12 (44.4\%) \\
Perception (Test) & 7 (18.4\%) & 3 (11.1\%) \\
Reasoning (Deductive) & 1 (2.6\%) & 2 (7.4\%) \\
\bottomrule
\end{tabularx}
\end{minipage}

\vspace{-1mm}

\begin{minipage}{0.48\linewidth}
\centering
\subcaption{ACRE}
\begin{tabularx}{\linewidth}{@{}lrr@{}}
\toprule
Setting & 3(a) & 3(b) \\
\midrule
\textbf{Total Errors} & 38  & 32  \\
Perception (Demo) & 29 (76.3\%) & 22 (68.8\%) \\
Reasoning (Inductive) & 6 (15.8\%) & 7 (21.9\%) \\
Perception (Test) & 3 (7.9\%) & 3 (9.4\%) \\
Reasoning (Deductive) & 0 (0\%) & 0 (0\%) \\
\bottomrule
\end{tabularx}
\end{minipage}
\hfill
\begin{minipage}{0.48\linewidth}
\centering
\subcaption{ACRE (varying P)}
\begin{tabularx}{\linewidth}{@{}lrr@{}}
\toprule
Setting & 3(b) & 3(c) \\
\midrule
\textbf{Total Errors} & 32  & 9  \\
Perception (Demo) & 22 (68.8\%) & 0 (0\%) \\
Reasoning (Inductive) & 7 (21.9\%) & 9 (100\%) \\
Perception (Test) & 3 (9.4\%) & 0 (0\%) \\
Reasoning (Deductive) & 0 (0\%) & 0 (0\%) \\
\bottomrule
\end{tabularx}
\end{minipage}
\vspace{-6mm}
\end{table*}

\begin{figure}[!tb]
\vspace{-6mm}
\centering

\setlength{\tabcolsep}{4pt}
\renewcommand{\arraystretch}{0.9}

\begin{subfigure}{0.46\columnwidth}
    \centering
    \includegraphics[width=1.05\linewidth]{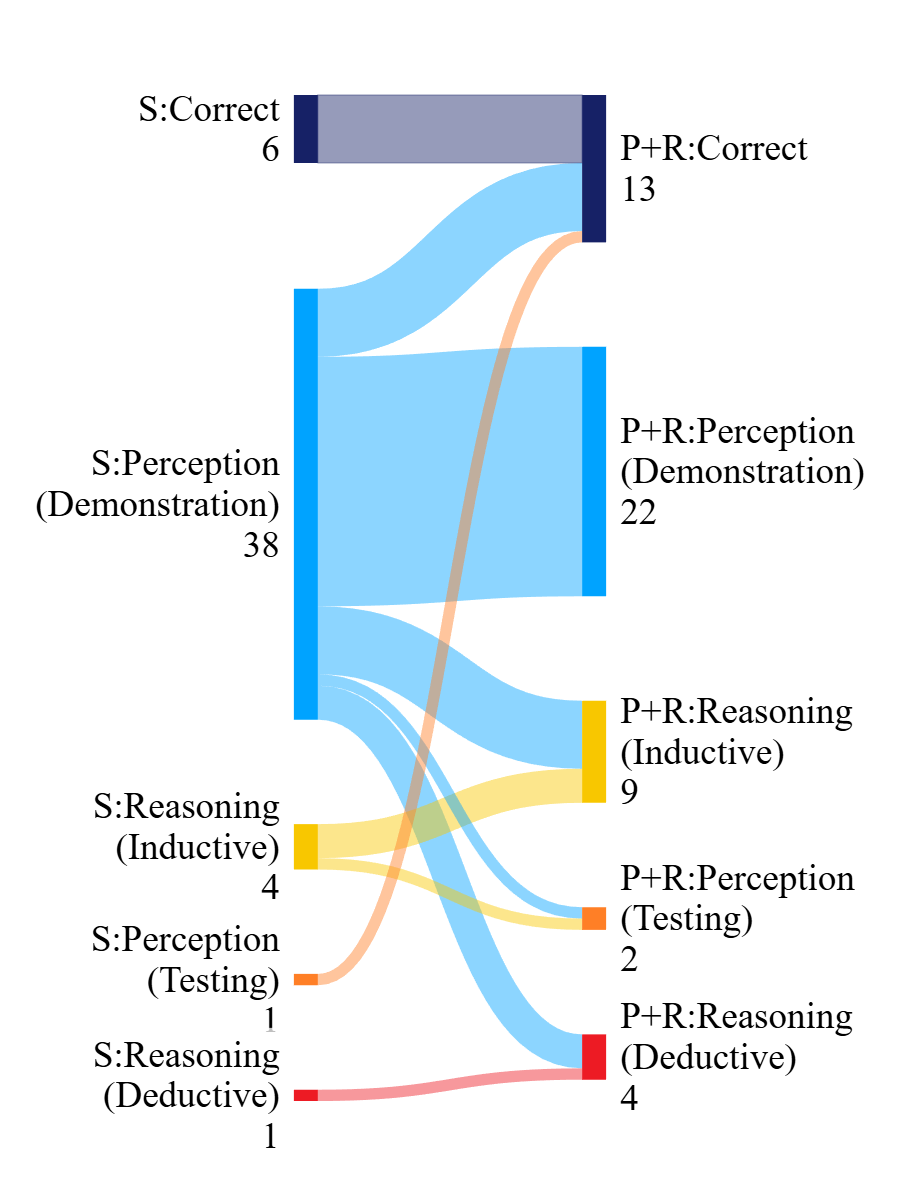}
    \caption{1(a) $\rightarrow$ 1(b)}
\end{subfigure}
\hfill
\begin{subfigure}{0.46\columnwidth}
    \centering
    \includegraphics[width=1.05\linewidth]{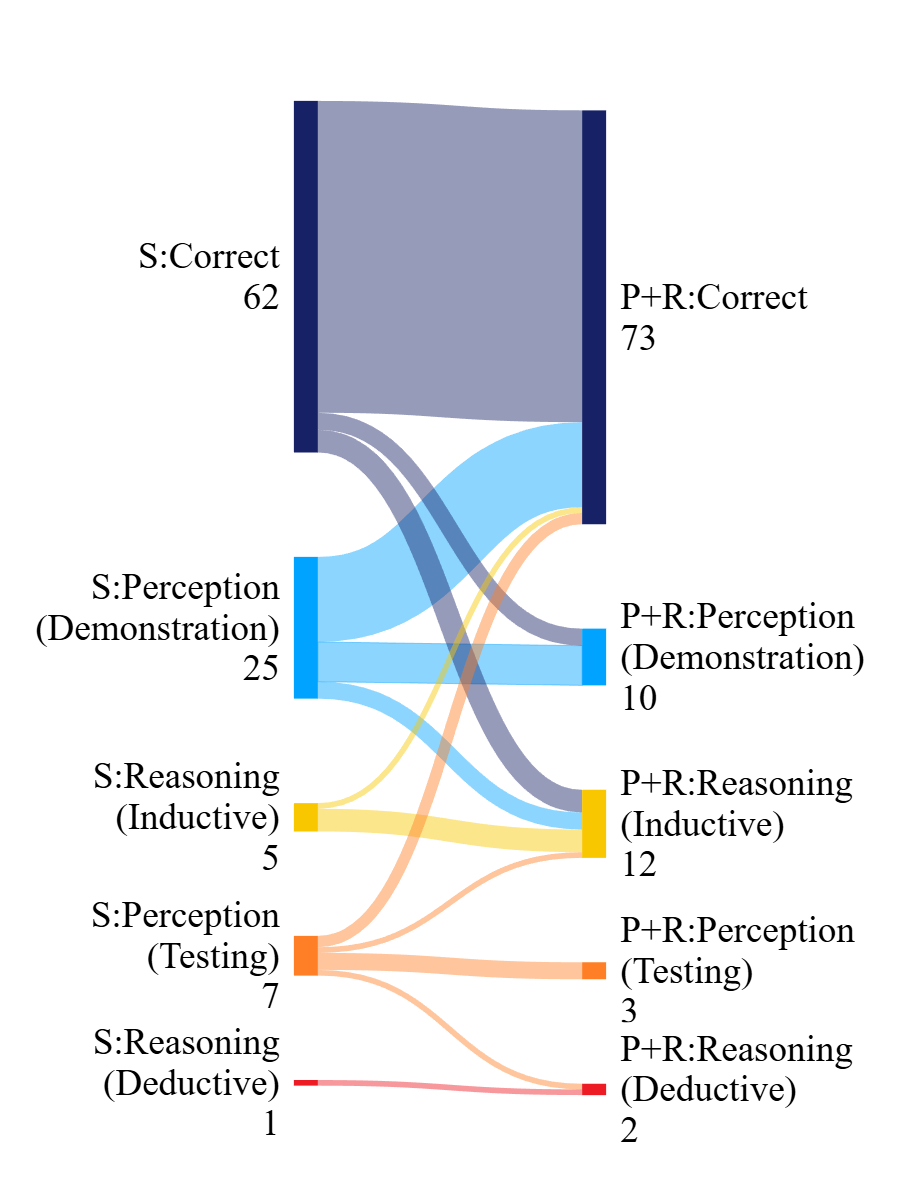}
    \caption{2(a) $\rightarrow$ 2(b)}
\end{subfigure}

\begin{subfigure}{0.46\columnwidth}
    \centering
    \includegraphics[width=1.05\linewidth]{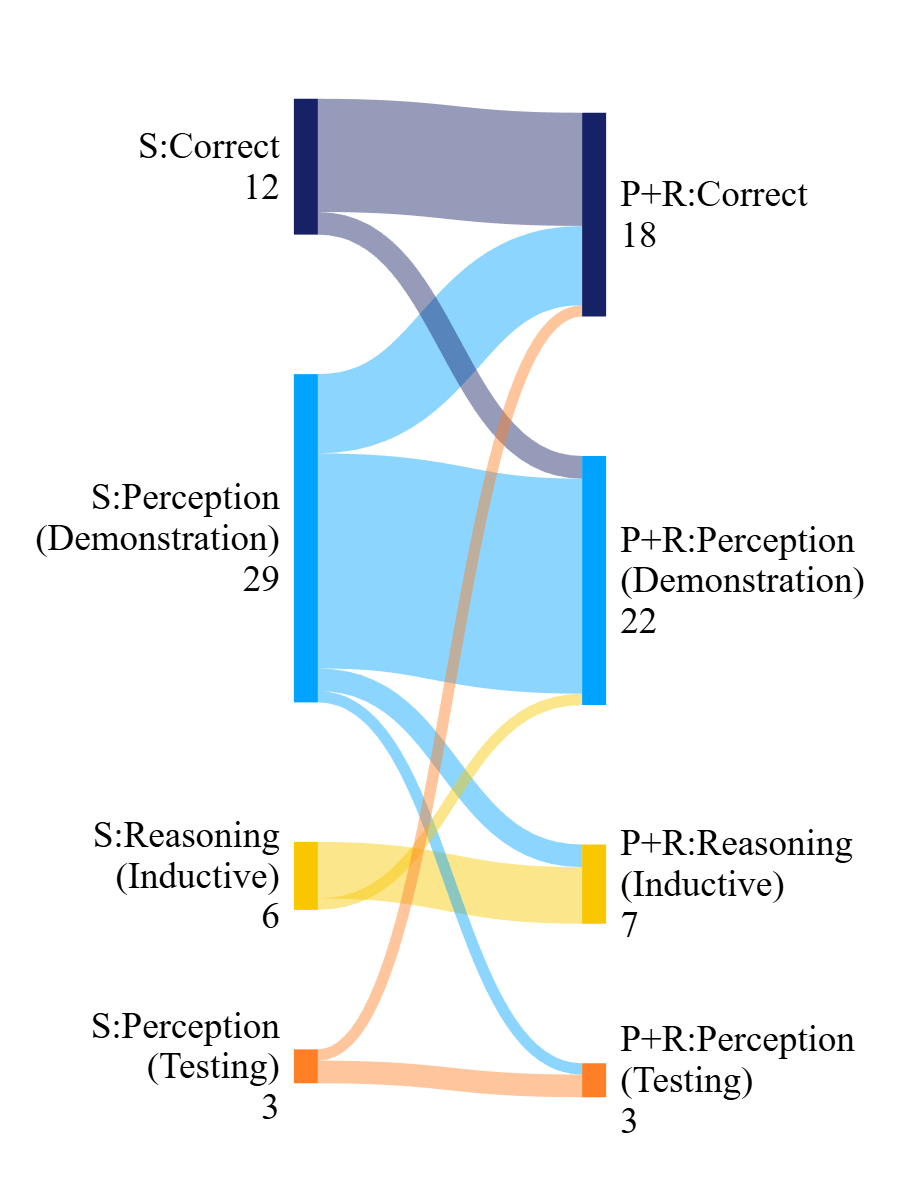}
    \caption{3(a) $\rightarrow$ 3(b)}
\end{subfigure}
\hfill
\begin{subfigure}{0.46\columnwidth}
    \centering
    \includegraphics[width=1.05\linewidth]{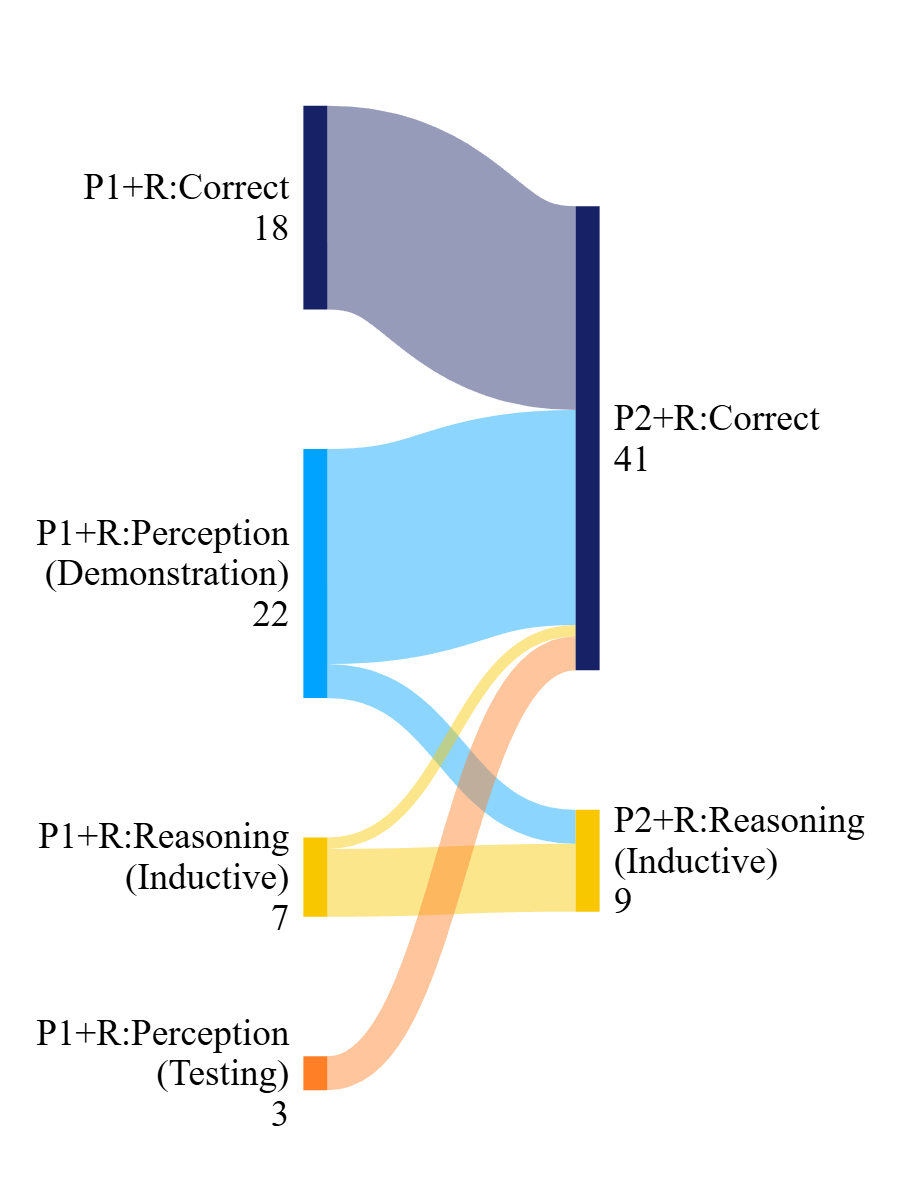}
    \caption{3(b) $\rightarrow$ 3(c)}
\end{subfigure}
\caption{
Visualization of how errors evolve from one configuration to another.
}
\label{fig:all-flow}
\vspace{-8mm}
\end{figure}
\paragraph{Results.}
As can be seen in Table~\ref{tab:setting2-results}, 
strengthening the perception module ((b) $\rightarrow$ (c*)\footnote{We use (c*) as a general reference to (c), (c1), and (c2). Same for (d*) below.}) yields consistent and substantial improvement, indicating that perception is indeed the dominant bottleneck in this setting. Furthermore, the performance in (c*) and that in (d*), the standard one-stage pipeline with the strong model, are mostly close\footnote{ One exception is Mini-ARC with o1, where the gap between (c1) and (d1) remains large. A plausible explanation is that Mini-ARC may have been included in o1’s training data. This possibility is supported by the observation that o1’s performance on Mini-ARC (d1) is markedly higher than that of Claude-Sonnet-4.5 (d2), while its performance on ARC-AGI is inferior to Claude-Sonnet-4.5.}, suggesting that perception constitutes the main bottleneck explaining the performance difference between the strong model (d*) and the weak model (a).

\subsection{Error Attribution}
\label{sec:error-attribution}
In order to further validate the hypothesis and gain insights into the causes of errors, we conduct error attribution on the failure cases for different configurations\footnote{We omitted configurations with o1 or Claude-Sonnet-4.5 as we did not have full reasoning traces in our results.} across datasets, following the protocol introduced in Section~\ref{sec:error-attribution-approach}. Specifically, for configurations on Mini-ARC and ACRE, we randomly select 50 tasks, and for those on Bongard-LOGO, we randomly select 100 tasks. Among these, some tasks may have been correctly solved by the model, while for the remaining ones we perform detailed error attribution. 

\paragraph{Results.} As shown in Table~\ref{tab:error-attrib-main}, perception errors consistently dominate across all setups, indicating that the visual understanding stage is the primary bottleneck. On Mini-ARC, perception errors account for 86.4\% of all errors in setting 1(a) (standard one-stage) and 59.5\% in setting 1(b) (two-stage). A similar trend is observed on Bongard-LOGO, where perception errors represent 65.8\% (2(a)) and 37.0\% (2(b)) of total errors, and on ACRE, where perception errors reach 76.3\% (3(a)) and 68.8\% (3(b)). On ACRE with varying P, where different perception modules are compared, the model with LLaVA1.5 still shows 68.8\% perception errors, while GPT-4o eliminates perception errors, with all remaining errors in reasoning stages.

These results clearly demonstrate that the perception stage remains the dominant source of model failure, overshadowing reasoning-related errors (both inductive and deductive). Moreover, this fine-grained error attribution provides more direct evidence that the performance improvement in both Setting 1 and Setting 2 are achieved through the mitigation of perception errors. This result highlights the importance of perception in tasks that are often regarded as reasoning tasks.

To gain deeper insight into this attribution, we further analyze how error types evolve across experimental configurations, specifically, how perception-related errors transition into reasoning errors (and vice versa) under different configurations, shown in Figure~\ref{fig:all-flow}. From the analysis, we observe that a substantial portion of perception errors are either eliminated or transformed into downstream reasoning errors, and that the majority of performance gains arise from resolving perception-related errors rather than improving reasoning.

\section{Conclusion}
In this work, we critically re-examine the interpretation of ARC-style benchmarks as direct measures of machine reasoning ability and show that a substantial portion of the observed human–AI performance gap is driven instead by limitations in visual perception. By introducing a carefully controlled two-stage pipeline that explicitly separates perception from reasoning, and by conducting fine-grained error attribution across multiple benchmarks, we demonstrate that mitigating perceptual bottlenecks alone leads to large performance gains, while improvements in reasoning play a secondary role. These findings suggest that ARC-style benchmarks conflate perceptual and inductive challenges, potentially overstating deficiencies in model reasoning. More broadly, our results underscore the importance of evaluation protocols that disentangle perception from reasoning when assessing progress toward general intelligence, and point to the need for future benchmarks that more cleanly isolate the cognitive capabilities they aim to measure.

\section*{Limitations}
Our study has several limitations. First, the natural-language descriptions used in the perception stage should not be interpreted as an optimal or canonical intermediate representation for abstract reasoning. We adopt language as a convenient and interpretable way to inject generic human perceptual priors and to demonstrate the existence and impact of perceptual bottlenecks in ARC-style benchmarks. We do not claim that natural language is the best representation for isolating reasoning, nor that it preserves all aspects of the original task structure without distortion.

Second, our fine-grained error attribution relies on manual inspection of model outputs and reasoning traces. While this attribution follows a clearly defined, stage-wise protocol and yields consistent patterns across datasets and experimental settings, it involves some subjectivity. We leave the development of scalable and fully automated attribution methods to future work.

Finally, our conclusions are limited to ARC-style, knowledge-light visual abstraction benchmarks and to the vision-language models evaluated in this study. We do not claim that perception is the dominant bottleneck for all multimodal or textual reasoning tasks. Nevertheless, we hope that the diagnostic perspective and methodology introduced in this work can inspire future research on disentangling perception and reasoning in a broader range of reasoning benchmarks and evaluation settings.

\bibliography{custom}

\end{document}